\documentclass[journal]{IEEEtran}
\usepackage[normalem]{ulem}
\usepackage{url}
\usepackage{cite}
\usepackage{balance}

\ifCLASSINFOpdf

  \usepackage[pdftex]{graphicx}
  \DeclareGraphicsExtensions{.pdf,.jpg,.png}
  \graphicspath{C:/Users/Amichai/Desktop/BICA transactions/}
\else
\fi
\usepackage{caption}
\usepackage{amssymb}
\usepackage[cmex10]{amsmath}

\usepackage{mathtools}
\usepackage{array}
\usepackage{algorithm}
\usepackage{algorithmic}
\usepackage{xfrac }

\newtheorem{theorem}{Theorem}

\newtheorem{appendix_lemma}{Lemma}

\newenvironment{proof}[1][Proof]{\begin{trivlist}
\item[\hskip \labelsep {\bfseries #1}]}{\end{trivlist}}

\newcommand{\qed}{\nobreak \ifvmode \relax \else
      \ifdim\lastskip<1.5em \hskip-\lastskip
      \hskip1.5em plus0em minus0.5em \fi \nobreak
      \vrule height0.75em width0.5em depth0.25em\fi}

\newcommand{\RNum}[1]{\uppercase\expandafter{\romannumeral #1\relax}}

\begin{document}
%
\title{Linear Independent Component Analysis\\ over Finite Fields: Algorithms and Bounds}
%
%
%

\author{Amichai~Painsky,~\IEEEmembership{Member,~IEEE,}
        Saharon~Rosset and~Meir~Feder,~\IEEEmembership{Fellow,~IEEE}

\thanks{A. Painsky is with the School of Computer Science and Engineering, The Hebrew University of Jerusalem, Israel. contact: amichai.painsky@huji.mail.ac.il}
\thanks{S. Rosset is with the Statistics Department, Tel Aviv University,  Israel.}
\thanks{M. Feder is with the Department of Electrical Engineering, Tel Aviv University,  Israel}
\thanks{The material in this paper was presented in part at the 2016 IEEE International Workshop on Machine Learning for Signal Processing}}

%
%

\markboth{IEEE TRANSACTIONS ON SIGNAL PROCESSING}%
{Shell \MakeLowercase{\textit{et al.}}: Bare Demo of IEEEtran.cls for Journals}
%



\maketitle

\begin{abstract}

Independent Component Analysis (ICA) is a statistical tool that decomposes an observed random vector into components that are as statistically independent as possible. ICA over finite fields is a special case of ICA, in which both the observations and the decomposed components take values over a finite alphabet. This problem is also known as \textit{minimal redundancy representation} or \textit{factorial coding}. In this work we focus on linear methods for ICA over finite fields. We introduce a basic lower bound which provides a fundamental limit to the ability of any linear solution to solve this problem. Based on this bound, we present a greedy algorithm that outperforms all currently known methods. Importantly, we show that the overhead of our suggested algorithm (compared with the lower bound) typically decreases, as the scale of the problem grows. In addition, we provide a sub-optimal variant of our suggested method that significantly reduces the computational complexity at a relatively small cost in performance. Finally, we discuss the universal abilities of linear transformations in decomposing random vectors, compared with existing non-linear solutions. 
\end{abstract}


%
\IEEEpeerreviewmaketitle

\section{Introduction}
%
%
%
%


\IEEEPARstart{I}{ndependent}  Component Analysis (ICA) addresses the recovery of unknown statistically independent source signals from their observed mixtures, without full prior knowledge of the mixing model or the statistics of the source signals. The classical Independent Components Analysis framework usually assumes linear combinations of the independent sources over the field of real valued numbers. A special variant of the ICA problem is when the sources, the mixing model and the observed signals are over a finite field, such as Galois Field of order $q$, GF($q$).

Several types of generative mixing models may be assumed when working over GF($q$). This includes modulo additive operations, OR operations (over a binary field) and others. Existing solutions to ICA mainly differ in their assumptions of the generative model, the prior distribution of the mixing matrix (if such exists) and the noise model. Nevertheless, the common assumption to all of these approaches is that there exists a set of fully statistically independent source signals to be decomposed. However,  these assumptions do not usually hold, and a more robust approach is required. In other words, we would like to decompose any given observed mixture into ``as independent as possible" components, with no prior assumption on the way it was generated. This problem was first introduced by Barlow \cite{barlow1989finding} as \textit{minimal redundancy representation} and was later referred to as \textit{factorial representation} \cite{schmidhuber1992learning} or \textit{generalized ICA over finite alphabets}\cite{painsky2016generalized}. 

A factorial representation has several advantages. The probability of occurrence of any realization can be simply computed as the product of the probabilities of the individual components that represent it (assuming such decomposition exists). In addition, any method of finding factorial codes can be viewed as implementing the \textit{Occam's razor} principle, which prefers simpler models over more complex ones, where simplicity may be defined as the number of parameters necessary to represent the joint distribution of the data.
In the context of supervised learning, independent features can also make later learning easier; if the input units to a supervised learning networks are uncorrelated, then the Hessian of its error function is diagonal, allowing accelerated learning abilities \cite{becker1988improving}. There exists a large body of work which demonstrates the use of factorial codes in learning problems. This mainly manifests in artificial neural networks \cite{becker1996unsupervised, obradovic1996information} with application to facial recognition \cite{choi2000factorial, bartlett2002face, bartlett2007information}  and  deep learning \cite{schmidhuber2011fast, schmidhuber2015deep}. Recently, factorial codes were also shown to attain favorable compression rates in large alphabet source coding  \cite{painsky2015universal,painsky2016simple,painsky2017large,painsky2018}.

In this work we focus on linear solutions to the factorial representation problem; we seek for a linear transformation (over GF($q$)), that decomposes the observed mixture into ``as statistically independent as possible" components. Linear solutions have several advantages. They are easy to implement in linear time-invariant (LTI)  systems, they are robust to noisy measurements and they are storage space efficient. Importantly, they are usually simpler to derive and analyze than in the non-linear case. For these reasons, linear ICA has received most of the attention over the years \cite{yeredor2007ica, yeredor2011independent,  vsingliar2006noisy, streich2009multi, nguyen2011binary, attux2011immune,wood2012non}. 

This paper is an extended version of our initial results, presented in \cite{painsky2016binary}. In \cite{painsky2016binary} we introduced a lower bound to the linear binary ICA problem, followed by a simple practical algorithm. Here we significantly extend the scope of this study. This includes the following contributions:
\begin{itemize}
\item A comprehensive derivation of the computational and statistical properties of the methods presented in \cite{painsky2016binary}.
\item  A novel computationally efficient algorithm which reduces the complexity of our previously suggested schemes.
\item  Generalization of our framework to finite fields of any order.
\item A rigorous study of the flexibility of linear ICA methods, including both analytical and empirical results.
\item Comparative study of linear and non-linear ICA methods. 
\item A data compression application, which demonstrates the use of our suggested approach in large alphabet source coding.  
\end{itemize}
\section{Previous Work}
\label{previous_work}
The problem considered in this paper has a long history. In his pioneering work from $1989$, Barlow \cite{barlow1989finding} presented a \textit{minimally redundant  representation} scheme for binary data. He claimed that a good representation should capture and remove the redundancy of the data. This leads to a \textit{factorial representation/ encoding} in which the components are as mutually independent of each other as possible. Barlow suggested that such representation may be achieved through \textit{minimum entropy encoding}: an invertible transformation (i.e., with no information loss) which minimizes the sum of marginal entropies (as later presented in (\ref{eq:min_criterion})). Unfortunately, Barlow did not propose any direct method for finding factorial codes. Atick and Redlich \cite{atick1990towards} proposed a cost function for Barlow's principle for linear systems, which minimize the redundancy of the data subject to a minimal information loss constraint. This is closely related to Plumbey's \cite{plumbley1993efficient}  objective function, which minimizes the information loss subject to a fixed redundancy constraint. Schmidhuber \cite{schmidhuber1992learning} then proposed several ways of approximating Barlow's minimum redundancy principle in the non-linear case. This naturally implies much stronger results of statistical independence. However, Schmidhuber's scheme is rather complex, and subjects to local minima \cite{becker1996unsupervised}. Recently,  we introduced a novel approach for finding factorial codes over non-linear transformation \cite{painsky2016generalized}. In that work, Barlow's objective is tightly approximated with a series of linear problems. Despite its favorable computational properties, the approach suggested in \cite{painsky2016generalized} is quite analytically involved. Later, we introduced a simpler sub-optimal solution, which applies an order permutation according to the probability mass function of the observed mixture \cite{painsky2017large}. This method is shown to minimize the desired objective function up to a small constant, for increasing alphabets sizes, on the average.  

In a second line of work, the factorial coding problem is studied under the assumptions that a fully independent decomposition exists, and that the mixture is linear. In his first contribution to this problem, Yeredor \cite{yeredor2007ica} considered a linear mixture of statistically independent sources over GF(2) (namely, \textit{binary ICA}) and proposed a method for source separation based on entropy minimization. Yeredor assumed that the mixing model is a $d$-by-$d$ invertible matrix and proved that the XOR model is invertible and that there exists a unique transformation matrix to recover the independent components. Yeredor further suggested several algorithms for the linear binary ICA (BICA) problem, including the AMERICA and the enhanced MEXICO algorithms. Further, Yeredor generalized his work \cite{yeredor2011independent} to address the ICA problem over Galois fields of any prime number. His ideas were analyzed and improved by Gutch et al. \cite{gutch2012ica}.  In \cite{attux2011immune}, Attux et al.  extended Yeredor's formulation for a more robust setup, in which the sources are not necessarily independent and presented a heuristic immune-inspired algorithm \cite{attux2011immune}, which was later  improved and generalized to any GF($q$) \cite{silva2014cobica}.

In addition to generative XOR model, there exist several alternative mixture assumptions. In \cite{vsingliar2006noisy} and  \cite{wood2012non} the authors considered a noise-OR model, where the probabilistic dependency between observable vectors and latent vectors is modeled via the noise-OR conditional distribution. 
Streith et al. \cite{streich2009multi} studied the binary ICA problem, where the observations are either drawn from a signal following OR mixtures or from a noise component. The key assumption made in this work is that the observations are conditionally independent given the model parameters (as opposed to the latent variables). This greatly reduces the computational complexity and makes the scheme amenable to an objective descent-based optimization solution. In \cite{nguyen2011binary}, Nguyen and Zheng considered OR mixtures and proposed a deterministic iterative algorithm to determine the distribution of the latent variables and the mixing matrix. 

\section{Linear Binary Independent Component Analysis}

We begin our presentation by discussing the simpler binary case. Throughout this manuscript we use the following standard notation: underlines denote vector quantities, while their respective components are written without underlines but with an index. 
The probability function of the vector $\underline{X}$ is denoted as  $P_{\underline{X}}\left(\b{x}\right)  \triangleq P(X_1 = x_1,\dots, X_d= x_d) $, while $H\left(\underline{X}\right)$ is the entropy of $\underline{X}$. This means  $H\left(\underline{X}\right)=-\sum_{\b{x}} P_{\underline{X}}\left(\b{x}\right) \log{P_{\underline{X}}\left(\b{x}\right)}$ where the $\log{}$ function denotes a logarithm of base $2$, unless stated otherwise. Further, we denote the binary entropy of the parameter $p$ as $h (p)=-p\log{p}-(1-p)\log{(1-p)}$.

\subsection{Problem Formulation}
\label{problem_formulation}
Let $\underline{X}\sim P_{\b{x}}\left(\b{x}\right)$ be a given random vector of dimension $d$. We are interested in an invertible transformation, $\underline{Y}=g(\underline{X})$, such that the components of $\underline{Y}$ are ``as statistically independent as possible". Notice that the common notion of ICA  is not limited to invertible transformations (hence $\underline{Y}$ and  $\underline{X}$ may be of different dimensions). However, in our work we focus on this setup as we would like $\underline{Y}=g(\underline{X})$ to be ``lossless" in the sense that we do not loss any information. Further motivation to this setup is discussed in \cite{barlow1989finding, schmidhuber1992learning}.    

We distinguish between linear and non-linear invertible transformations. 
In the linear case, $g(\cdot)$ is a $d$-by-$d$ invertible matrix over the XOR field. This means we seek  $\underline{Y}=W\cdot\underline{X}$ where $W \in \{0,1\}^{(d \times d)}$ and $\text{rank}(W)=d$. In the non-linear case, we notice that an invertible transformation of a vector $\underline{X}$ is a one-to-one mapping (i.e., permutation) of its $2^d$ words. This means there exist $2^d!$ invertible transformations from which only $O\left( 2^{d^2} \right)$ are linear \cite{yeredor2011independent}.

To quantify the statistical independence among the components of $\underline{Y}$ we use the well-known \textit{total correlation} criterion, which was first introduced by Watanabe \cite{watanabe1960information} as a multivariate generalization of the mutual information,
\begin{equation}\label{eq:min_criterion}
C\left(\underline{Y}\right)={\displaystyle \sum_{j=1}^{d}{H(Y_j)}-H(\underline{Y})}.
\end{equation}
This measure can also be viewed as the cost of coding the vector $\underline{Y}$ component-wise, as if its components were statistically independent, compared to its true entropy. Notice that the total correlation is non-negative and equals zero iff the components of $\underline{Y}$ are mutually independent. Therefore, ``as statistically independent as possible" may be quantified by minimizing $C(\underline{Y})$. The total correlation measure was first considered as an objective for factorial representation by Barlow \cite{barlow1989finding}. 

Since we define $\underline{Y}$ to be an invertible transformation of $\underline{X}$ we have $H(\underline{Y})=H(\underline{X})$ and so our minimization objective, in the binary case, is
\begin{equation}
{\displaystyle \sum_{j=1}^{d}{H(Y_j)}=\sum_{j=1}^{d}{h(P(Y_j=0))}  \rightarrow min,}
\label{eq:sum_ent_min_binary}
\end{equation}
where $P(Y_j=0)$ is the sum of probabilities of all words whose $j^{th}$ bit equals $0$. This means that the transformations are not unique. For example, we can invert the  $j^{th}$ bit of all words to achieve the same objective, or even shuffle the bits.

Notice that the probability mass function of $\underline{X}$ is defined over $2^d$ values.  Therefore, any approach which exploits the full statistical description of $\underline{X}$ would require going over all $2^d$ possible words at least once.  On the other hand, there exist at most  $2^d!$ possible invertible transformations. The complexity of currently known binary ICA methods (and factorial codes)  fall within this range. In the linear case (and under a complete independence assumption), the AMERICA algorithm \cite{yeredor2011independent}, which assumes a XOR mixture, has a complexity of $O\left(d^2 2^d\right)$. The MEXICO algorithm, which is an enhanced version of AMERICA, achieves a complexity of $O\left(2^d\right)$ under some restrictive assumptions on the mixing matrix. In the non-linear case, an approximation of the optimal solution may be achieved in a computational complexity of $O\left(d^k 2^d\right)$, where $k$ is the accuracy parameter \cite{painsky2014generalized}, while the simpler order permutation \cite{painsky2017large} requires $O\left(d2^d\right)$.

\subsection{Lower Bound on Linear Binary ICA}
\label{linear_BICA_lowerbound}

In this section we introduce a lower bound to the binary linear ICA problem.  Specifically, we present a method which obtains an infimum of  (\ref{eq:sum_ent_min_binary}), under $\underline{Y}=W\underline{X}$ and $W \in \{0,1\}^{(d \times d)}$, $\text{rank}(W)=d$. 

In his linear binary ICA work, Yeredor established a methodology based on a basic property of the binary entropy \cite{yeredor2011independent}. He suggested that the binary entropy of the XOR of two independent binary variables is greater than each variable's entropy. 
Unfortunately, there is no such guarantee when the variables are dependent. This means that in general, the entropy of the XOR of binary variables may or may not be greater than the entropy of each of the variables (for example, $H(X\oplus X)=0$, where $\oplus$ is the xor operand). When minimizing  (\ref{eq:sum_ent_min_binary}) over linear transformations,  $\underline{Y}=W\underline{X}$, we notice that each $Y_j$ is a XOR of several, possibly dependent, variables $X_1,\dots,X_d$. This means that naively, we need to go over all possible subsets of  $\{X_1,\dots,X_d\}$ and evaluate their XOR. Formally, we define a matrix $A$ of all possible $2^d$ realizations of $d$ bits. Then, we would like to compute $U_i= A_{i1}X_1 \oplus A_{i2}X_2\oplus \ldots \oplus A_{id}X_d$ for all $i=1,\dots,2^d$. This means that each row in the matrix $A$ corresponds to a subset of variables from $\{X_1,\dots,X_d\}$, on which we apply a XOR. Then, we evaluate the binary entropy of each $U_i$. A necessary condition for $W$ to be invertible is that it has no two identical rows. Therefore, a lower bound on  (\ref{eq:sum_ent_min_binary}) may be achieved by simply choosing $d$ rows of the matrix $A$, for which $H(U_i)$ are minimal. Notice that this lower bound is not tight or attainable. It defines a simple lower bound on (\ref{eq:sum_ent_min_binary}), which may be attained if we are lucky enough to have chosen $d$ rows of the matrix $A$ which are linearly independent.     

The derivation of our suggested bound requires the computation of $2^d$ 
binary entropy values, where each entropy corresponds to a different vectorial XOR operation of length $d$. This leads to a computational complexity of $O\left(d 2^d\right)$. Then, we sort the $2^d$ entropy values in an ascending order, using a simple quick-sort implementation. This again requires $O\left(d 2^d\right)$ operations. Therefore, the  total computation complexity of our suggested bound is  $O\left(d 2^d\right)$. In other words, we attain a lower bound to any linear solution of (\ref{eq:sum_ent_min_binary}), in a computational complexity that is asymptotically competitive to all currently known methods (for both linear and non-linear methods).

\subsection{A Greedy Algorithm for Linear Binary ICA}
\label{linear_BICA_algo}
We now present our suggested algorithmic approach for the linear BICA problem, based on the same methodology presented in the previous section.  
Again, we begin by evaluating all possible XOR operations $U_i= A_{i1}X_1 \oplus A_{i2}X_2\oplus \ldots \oplus A_{id}X_d$ for all $i=1,\dots,2^d$. Further, we evaluate the binary entropy of each $U_i$. We then sort the rows of $A$ according to the binary entropy values of their corresponding $U_i$.  
Denote the sorted list of rows as $\tilde{A}$. Our remaining challenge is to choose $d$ rows from $\tilde{A}$ such that the rank of these rows is $d$. Clearly, our objective suggests that we choose rows which are located at the top of the matrix $\tilde{A}$, as they result in lower entropy values. Our suggested greedy algorithm begins with an empty matrix $W$. It then goes over the rows in $\tilde{A}$ in an ascending order. If the current row in $\tilde{A}$ is linearly independent of the current rows in $W$ it is added to $W$. Otherwise, it is skipped and the algorithm proceeds to the next row in $\tilde{A}$. The algorithm terminates once $W$ is of full rank. The rank of $W$ is evaluated by a simple Gauss elimination procedure (implemented on a Matlab platform as $\sf{gfrank}$, for example) or by more efficient parallel computation techniques \cite{mulmuley1986fast}.
 We denote our suggested algorithm as \textit{Greedy Linear ICA} (GLICA). 

Although GLICA looks for $d$ linearly independent rows from $\tilde{A}$ in a no-regret manner, we may still consider its statistical properties and evaluate how much it deviates from the lower bound. Let us analyze the case where the order of the rows is considered random. This means that although 
the rows of $\tilde{A}$ are sorted according to the value of their corresponding entropy, $H(U_i)$, we still consider the rows as randomly ordered, in the sense that the position of each row in $\tilde{A}$ is random and independent of the other rows. This assumption is typically difficult to justify. However,  we later demonstrate in a series of experiments that the derived properties may explain the favorable performance of GLICA. This suggests that although there is typically dependence among the row, it is practically low.


Let us examine GLICA at an iteration where it already found $k<d$ linearly independent rows (rank$(W)=k$), and it is now seeking an additional independent row. Notice that there exist at most  $2^k$ rows which are linearly dependent of the current rows in $W$ (all possible linear combinations of these rows). Assume we uniformly draw (with replacement) an additional row from a list of all possible rows (of size $2^d$). The probability of a drawn row to be independent of the $k$ rows in $W$ is simply $1-\sfrac{2^k}{2^d}$. Therefore, the number of draws needed in order to find another linearly interdependent row follows a geometric distribution with a parameter $1-\sfrac{2^k}{2^d}$ (as the draws are i.i.d.). This means that the average number of draws needed to find an additional row is $\frac{1}{1-\sfrac{2^k}{2^d}}=\frac{2^d}{2^d-2^k}$, while its variance is $\frac{2^{d+k}}{(2^d-2^k)^2}$. Denote the number of rows that GLICA examines before termination by $L$. We have that
$\mathbb{E}(L)=\sum_{k=0}^{d-1}\frac{2^d}{2^d-2^k}$ and  $\text{var}(L)\leq 2^{d} \sum_{k=0}^{d-1}\frac{2^{k}}{(2^d-2^k)^2}$. It can be shown \cite{shulman2003communication} that  $\mathbb{E}(L)-d \leq 2$ and $\lim_{d \rightarrow \infty} \mathbb{E}(L)-d = 1.606$. Further, $\textit{var}(L) \leq 2.744$. We may now apply Chebyshev's inequality to conclude that for every $a>0$ we have that 
\begin{align}
&P\left( L \geq d+2+a \right) =\\\nonumber
& P\left(L-\mathbb{E}(L)+\mathbb{E}(L)-(d+2) \geq a \right) \leq \\\nonumber 
&P\left( \left|L-\mathbb{E}(L)\right| \geq a \right)  \leq  \frac{\text{var}(L)}{a^2}\leq \frac{2.744}{a^2}. 
\end{align}
This result implies that if we choose rows from $\tilde{A}$, even with replacement, our suggested algorithm skips up to $2$ rows on the average, before terminating with a full rank matrix $W$, under the assumptions mentioned above. Further, the probability that our algorithm skips $2+a$ rows ($a>0$) is bounded from above by $\frac{2.744}{a^2}$. Notice that this bound is independent of $d$. Therefore, the overhead from our suggested the lower bound becomes negligible, as $d$ increases. For example, for $d=100$, the probability that we will examine $108$ rows or more ($a=6$) is not greater than $0.077$.

\subsection{Block-wise Approach for Linear Binary ICA}
\label{BloGLICA}
Despite its favorable computational and statistical properties, our suggested greedy algorithm (GLICA) still requires the computation of all possible $2^d$ XOR operations. This becomes quite costly as $d$ increases (or as alphabet size grows, as we see in Section \ref{Linear ICA over Finite Fields}). Therefore, we suggest a simple modification to circumvent this computational burden. 

Let us split the $d$ components of $\underline{X}$ into $b$ disjoint sets, $\{B_j\}_{j=1}^b$ (blocks).  For example, for $d=7$ and $b=3$ we may have that $X_1,X_2 \in B_1$ , $X_3,X_4 \in B_2$ and $X_5,X_6, X_7 \in B_3$. Let us now apply  GLICA to each block. Denote the outcome of this operation as $\underline{Y}=W_b\underline{X}$, where the matrix $W_b$ is block diagonal. As discussed in Section \ref{problem_formulation}, our objective (\ref{eq:sum_ent_min_binary}) is invariant to shuffling of components. It means we can randomly shuffle the components of $\underline{Y}$ (by multiplying it with a permutation matrix), and maintain the same objective value $\sum_{j=1}^d H(Y_j)$. In our example ($d=7$, $b=3$), the random shuffling may attain, for example, $Y_5,Y_1 \in B_1$ , $Y_2,Y_7 \in B_2$ and $Y_3,Y_4, X_6 \in B_3$. Notice that we now have a new set of blocks, on which we may again apply the GLICA algorithm, to further reduce our objective. We repeat this process for a configurable number of times, or until (local) convergence occurs. Notice that the convergence is guaranteed since we generate a sequence of non-increasing objective values, which is bounded from below by our suggested lower bound (Section \ref{linear_BICA_lowerbound}). The resulting linear transformation of this entire process is simply the multiplication of all  matrices that we apply along the process. We denote this iterative block-wise algorithm as \textit{Block GLICA},  or simply BloGLICA. Our suggested approached is summarized in Algorithm \ref{alg}

\begin{algorithm}[H] 
\caption{BloGLICA}
\begin{algorithmic} [1]
\REQUIRE $\underline{X}=\{X_j\}_{j=1}^d$, $b=$ the number of blocks, $M=$ maximal number of iterations. 
\STATE Set $W=I_{d\times d}$, a unit matrix. 
\STATE Set $\underline{Y}=\underline{X}$. 
\STATE Split $\{Y_j\}_{j=1}^d$ into $b$ disjoint blocks. \label{begin} 
\STATE Apply GLICA to each block to attain $\underline{Y}=W_b \underline{X}$ \label{generic}
\STATE Apply $\underline{Y}=U\underline{Y}$ where $U$ is a permutation matrix
\STATE  Set $W=W\cdot W_b \cdot U$ \label{end} 
\STATE Repeat steps  \ref{begin}-\ref{end}  $M$ times, or until convergence occurs.  
\RETURN $\underline{Y}, W$.
\end{algorithmic}
\label{alg}
\end{algorithm}

Notice that this block-wise approach is a conceptual framework which reduces the computational complexity of any finite alphabet ICA method. In other words, we may replace the GLICA algorithm in line \ref{generic} of Algorithm \ref{alg} by any finite field ICA algorithm, and by that reduce its computational burden. 
Let us set the maximal number of iterations to $M$. Further, set the maximal size of each block as $d_b=\lceil \frac{d}{b}\rceil$. Then, the computational complexity of BloGLICA is simply $O\left(M d_b 2^{d_b}\right)$. Notice this complexity is typically much smaller than GLICA's $O\left(d 2^{d}\right)$, since we do not examine all possible XOR operations and focus on local random searches that are implied by the block structure that we define.

\subsection{Illustrations}
\label{illustrations}
\begin{figure}[h]
\centering
\includegraphics[width = 0.45\textwidth,bb= 50 193 560 590,clip]{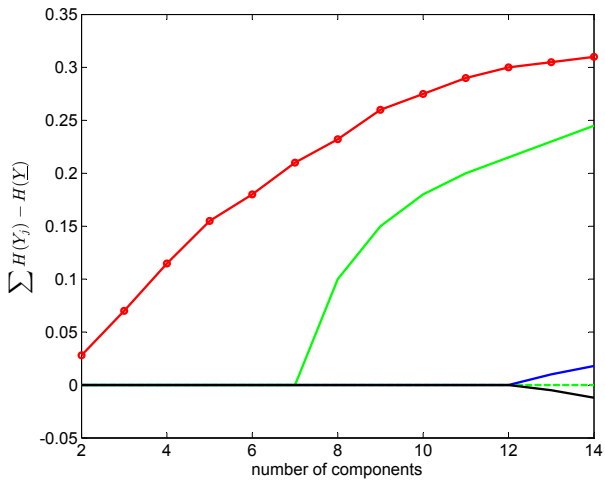}
\caption{Recovering independent sources experiment. Black curve at the bottom: lower bound on linear ICA. Blue curve: GLICA. Green dashed curve: AMERICA. Green curve: MEXICO. Red curve with the circles (on top): cobICA.}
\label{independent_sources_experiment}
\end{figure}

Let us now illustrate the performance of our suggested algorithms and bounds in several experiments. In the first experiment we examine our ability to recover $d$ independent binary sources that were mixed by an unknown matrix $B$. Let $\underline{S} \in \{0,1\}^d$ be a d-dimensional binary vector. Assume that the components of $\underline{S}$ are i.i.d. with a parameter  $p=0.4$. This means that the joint entropy of  $\underline{S}$ is $d\cdot h(0.4)$. We draw $10,000$ i.i.d. samples from $\underline{S}$ and mix them with a binary matrix $B$. Then, we  apply our binary ICA approach to recover the original samples and the mixing matrix $B$. Figure \ref{independent_sources_experiment} demonstrates the averaged results we achieve for different number of components $d$, where $B$ is an arbitrary invertible matrix that is randomly drawn, prior to the experiment. We verify that matrix $B$ is not an identity, nor a permutation matrix, to make the experiment meaningful. We compare our suggested approach with three alternative methods: AMERICA, MEXICO (described in Section \ref{previous_work}) and cobICA \cite{silva2014cobica}. As previously described, the cobICA is an immune-inspired algorithm. It starts with a random ``population" where each element in the population represents a valid transformation (an invertible matrix). At each step, the algorithm evaluates the objective function for each element in the population, and subsequently ``clones" the elements. Then, the clones suffer a mutation process, generating a new set of individuals. This new set is evaluated again in order to select the individuals with the best objective values. The entire process is repeated until a pre-configured  number of repetitions is executed. The cobICA algorithm requires a careful tuning of several parameters. Here, and in the following experiments, we follow the guidelines of the authors;  we set the \textit{general parameters} as appear in Table $1$ of  \cite{silva2014cobica}, while the rest of the parameters (\textit{concentration and suppression}) are then optimized over a predefined set of values, in a preliminary independent experiment.

We first notice that for $d \leq 12$, both GLICA and AMERICA successfully recover the mixing matrix $B$ (up to permutation of the sources), as they achieve an empirical sum of marginal entropies, which equals to the entropy of the samples prior to the mixture (blue curve at the bottom). Second, we notice that for the same values of $d$, our suggested lower bound (Section \ref{linear_BICA_lowerbound}) seems tight, as GLICA and AMERICA attain it. This is quite surprising since our bound is not expected to be attainable. This phenomenon may be explained by the simple setting and the relatively small dimension.  In fact, we notice that as the dimension increases beyond $d=12$, our suggested bound drops beneath the joint entropy, $H(\underline{Y})$, and GLICA fails to perfectly recover $B$.  The green curve corresponds to MEXICO, which demonstrates inferior performance, due to its design assumptions (see Section \ref{problem_formulation}) . Finally, the red curve with the circles is cobICA, which is less competitive as the dimension increases.  It is important to emphasize that while AMERICA and MEXICO are designed under the assumption that a perfect decomposition exists, cobICA and GLICA do not assume a specific generative model. Nevertheless, GLICA demonstrates competitive results, even when the dimension of the problem increases. 

\begin{figure}[h]
\centering
\includegraphics[width = 0.45\textwidth,bb= 50 193 550 590,clip]{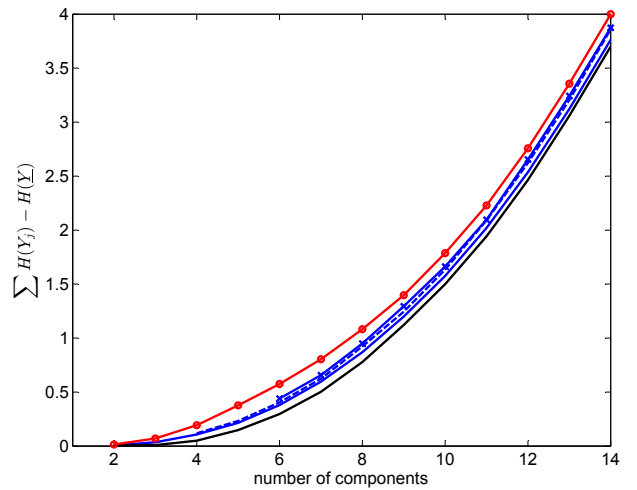}
\caption{Minimizing (\ref{eq:sum_ent_min_binary}) given independent draws from a Zipf's distribution. Black curve at the bottom: lower bound on linear ICA. Blue curve above it: GLICA. Blue dashed curve: BloGLICA with $b=2$. Blue curve with X's: BloGLICA with $b=3$. Red curve with circles (on top): cobICA.}
\label{fig1}
\end{figure}

In our second experiment we consider a binary source vector $\underline{X} \sim \underline{p}$ over an alphabet size $m=2^d$, whose joint distribution follows a Zipf's law distribution, 
\begin{equation}\nonumber
P(k;s,m)=\frac{k^{-s}}{\sum_{l=1}^m l^{-s}}
\end{equation}
where $m$ is the alphabet size and $s$ is the ``skewness" parameter. The Zipf's law distribution is a commonly used heavy-tailed distribution, mostly in modeling of natural (real-world) quantities. It is widely used in physical and social sciences, linguistics, economics and many other fields.  In this experiment, we draw $10,000$ i.i.d. random samples from a Zipf's law distribution with $s=1.01$, where each sample is represented by a $d$-dimensional binary vector, and the representation is chosen at random.
We evaluate the lower bound of (\ref{eq:sum_ent_min_binary}) under linear transformations, followed by the  GLICA algorithm. In addition, we examine the block-wise approach, BloGLICA (Section \ref{BloGLICA}) with $b=2, 3$.  We compare our suggested  bound and algorithms to cobICA \cite{silva2014cobica}.  Notice that in this experiment we omit the AMERICA and MEXICO algorithms, as they assume a generative mixture model, which is heavily violated in our setup. Figure \ref{fig1} demonstrates our objective (\ref{eq:sum_ent_min_binary}), for an increasing number of components $d$. 

\begin{figure}[h]
\centering
\includegraphics[width = 0.42\textwidth,bb= 50 193 550 590,clip]{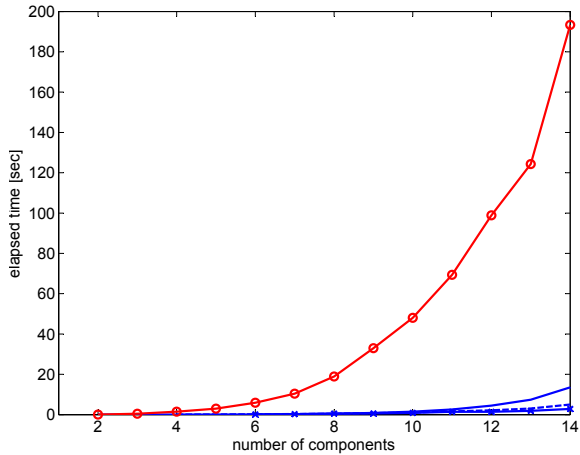}
\caption{Runtime of the experiment presented in Figure \ref{fig1}. 
}
\label{fig2}
\end{figure}

We first notice that our suggested algorithms lie between the lower bound and cobICA. Interestingly, bloGLICA preforms reasonably well, even as the number of blocks increases. On the computational side, Figure \ref{fig2} demonstrates the runtime of each of these methods, over a standard personal computer.  While we do not claim that these are the optimal implementations of the algorithms, obvious optimizations of the algorithms were implemented. The fact that they all were implemented in the same language (Matlab) is assumed to give none of the methods a significant advantage over the others. The same run-time comparison approach was taken, for example, in \cite{gutch2012ica}.

Here we notice the extensive runtime required by cobICA, compared to our suggested methods. In addition, we see that as the dimension increases, GLICA necessitates a rapidly increasing runtime (as it requires to compute and sort $2^d$ entropy values). However, by introducing BloGLICA we avoid this problem at a relatively small overhead in the objective.

\begin{figure}[h]
\centering
\includegraphics[width = 0.42\textwidth,bb= 50 193 550 590,clip]{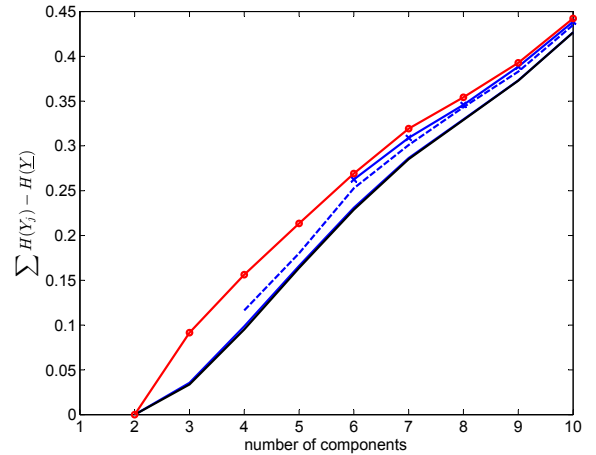}
\caption{Minimizing (\ref{eq:sum_ent_min_binary}) given independent draws from a Beta-Binomial distribution with $a=b=3$. The curves represent the same schemes described in Figure \ref{fig1}.}
\label{beta_binomial_experiment}
\end{figure}

Finally, we repeat the previous experiment, where the samples are now drawn from a source distribution with a greater entropy. Specifically, we consider $10,000$ i.i.d. draws from a Beta-Binomial distribution over an alphabet size $m=2^d$,  
\begin{equation}\nonumber
P(k;a,b,m)={m \choose k}\frac{\text{B}(k+a,m-k+b)}{\text{B}(a,b)}
\end{equation}
where $a$ and $b$ are the parameters of the distribution and $\text{B}(\cdot,\cdot)$ is the Beta function. We set $a=b=3$ in our experiment. The Beta-Binomial distribution is the Binomial distribution in which the probability of success at each trial is not fixed but random and follows the Beta distribution. It is frequently used in Bayesian statistics, empirical Bayes methods and classical statistics, to capture overdispersion in Binomial type distributed data. Although the entropy of this distribution does not hold an analytical expression, it can be shown to be greater than $d-\frac{1}{2}$ for $a=b=3$. As before, each drawn sample is represented by a $d$-dimensional binary vector, where the initial representation is chosen at random. Figure \ref{beta_binomial_experiment} demonstrates the results we achieve, applying the same binary ICA schemes as in the previous experiment. 

Here again, we notice the same qualitative behavior as with the Zipf's law experiment, where the main difference is that the ICA algorithms attain smaller values of (\ref{eq:sum_ent_min_binary}). The reason for this phenomenon lies on the observation that sources with greater entropy are typically easier to decompose into independent components. This property is further discussed in Section \ref{on_the_flexibility}.

\section{Linear ICA over Finite Fields}
\label{Linear ICA over Finite Fields}
Let us now extend our scope to general (possibly non-binary) finite fields of higher order. 
Finite fields (Galois fields) with $q$ elements are commonly denoted as GF($q$). A finite field only exists if $q$ is a prime power i.e. $q=p^z$ for some prime $p$ and a positive integer $z$. When , $z=1$ the field is called a \textit{prime field} and its elements can be represented by the integers $\{0, \dots,  ,q-1\}$. Addition, subtraction and multiplication are then performed by modulo-$q$ operations, while division is not completely straight-forward, but can be easily implemented using B\'ezout's identity \cite{gutch2012ica}. The simplicity of this field structure allows an easy implementation on any mathematical software -- all it requires is multiplication, addition and the modulo operation on integers. For these reasons, we focus on linear ICA over prime fields, although our results can be easily extended to any finite field of prime power. As with the binary case, we are interested in minimizing the sum of marginal entropies, where it is now defined as
\begin{equation}
\label{eq:sum_ent_min_prime}
\sum_{j=1}^dH(Y_j)=-\sum_{j=1}^d \sum_{a=0}^{q-1} P(Y_j=a)\log P(Y_j=a).
\end{equation}
Here, we have that $\underline{Y}=W\underline{X}$ where $W\in\{0,\dots,q-1\}^{d\times d}$ and the multiplication is modulo--$q$. 
As before, each $Y_j$ is a linear combination (over GF($q$)) of possibly dependent variables $X_1,\dots,X_d$. This means that naively, we need to go over all possible linear combinations of  $\{X_1,\dots,X_d\}$ to find the linear combinations that minimize the marginal entropies. Formally, we may define a matrix $D$ of all possible $q^d$ linear combinations. We would like to compute $U_i= \left(D_{i1}X_1 + D_{i2}X_2+ \ldots + D_{id}X_d\right)$$\mod q$,   for all $i=1,\dots,q^d$. Then, we evaluate the  entropy of each $U_i$. As before, a necessary condition for $W$ to be invertible is that it has no two identical rows. Therefore, a lower bound on  (\ref{eq:sum_ent_min_prime}) may be achieved by simply choosing $d$ rows of the matrix $D$, for which $H(U_i)$ are minimal. As in the binary case, this lower bound is not tight or attainable; it defines a simple bound on (\ref{eq:sum_ent_min_prime}), which may be attained if we are lucky enough to have chosen $d$ rows of the matrix $D$ that are linearly independent.  Notice that this bound requires the computation of $q^d$ entropy values, followed by sorting them in a ascending order. This leads to a computational complexity of $O\left(d q^d\right)$, which may become quite costly as $q$ and $b$ increase. 

As discussed in Section \ref{linear_BICA_algo}, we may derive a simple algorithm from our suggested bound, that seeks for $d$ independent rows from in a greedy manner. As in the binary case, we may derive the second order statistics of our suggested scheme, under the same assumptions mentioned in Section \ref{linear_BICA_algo}. Here, we have that the expected number of rows that our suggested algorithm examines before termination, $\mathbb{E}(L)$,  satisfies
\begin{align}
\label{expectation GP($q$)}
\mathbb{E}(L)-d=&\sum_{k=0}^{d-1}\frac{q^d}{q^d-q^k}-d=\sum_{k=0}^{d-1}\frac{1}{q^{d-k}-1}=\\\nonumber
&\sum_{k=1}^{d}\frac{1}{q^{k}-1}\leq \sum_{k=1}^{d}\frac{1}{q^{k}-q^{k-1}}\leq \frac{q}{(q-1)^2}.
\end{align}
Further, the variance is bounded from above by
\begin{align}
&\textit{var}(L)=\sum_{k=0}^{d-1}\frac{q^{d+k}}{\left(q^d-q^k\right)^2}=\sum_{k=1}^{d}\frac{1}{q^{k}+q^{-k}-2} \leq\\\nonumber
& \sum_{k=1}^{d}\frac{1}{q^{k}+q^{-k}-q^{k-1}}-\frac{1}{q+q^{-1}-1}+\frac{1}{q+q^{-1}-2}\leq \\\nonumber
&\frac{q}{(q-1)^2}+\frac{1}{\left(q+q^{-1}-2\right)^2.}
\end{align}
where the last inequality is due to (\ref{expectation GP($q$)}). A similar derivation is given in \cite{shulman2003communication}. Notice that both the expected overhead $\mathbb{E}(L)-d$ and the variance of $L$ converge to zero as the order of the field grows. 
We now apply Chebyshev's inequality to conclude that for every $a>0$ we have that 
\begin{align}
\label{cheby}
P\Bigg( L \geq d+ \frac{q}{(q-1)^2}+&a \Bigg) \leq \\\nonumber
&\frac{1}{a^2}\left(\frac{q}{(q-1)^2}+\frac{1}{\left(q+q^{-1}-2\right)^2}\right) .
\end{align}
Again, this result implies that under the independent draws assumption, choosing rows from the sorted ${D}$, even with replacement, skips up to $\frac{q}{(q-1)^2}$ rows on the average, before terminating with a full rank matrix $W$. Further, the probability that our algorithm will skip more than $\frac{q}{(q-1)^2}$ rows is governed by the left term of (\ref{cheby}). 
Unfortunately, despite these desired statistical properties, the computational complexity of this algorithm, $O\left(d q^d\right)$, becomes quite intractable as both $q$ and $d$ increase. Therefore, we  again suggest a block-oriented algorithm, in the same manner described in Section \ref{BloGLICA}. This reduces the computation complexity to 
 $O\left(M d_b q^{d_b}\right)$, where $M$ is the maximal number of iterations and $d_b=\lceil \frac{d}{b}\rceil$ is the maximal number of components in each of the $b$ blocks.

\begin{figure}[h]
\centering
\includegraphics[width = 0.42\textwidth,bb= 50 193 550 590,clip]{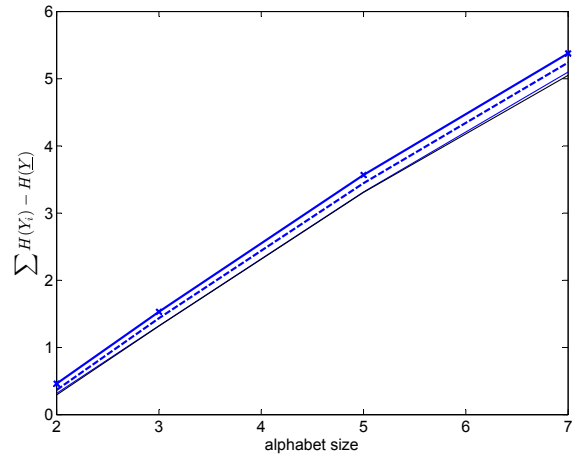}
\caption{Minimizing (\ref{eq:sum_ent_min_prime}) for $\underline{X} \in \{0,\dots, q-1\}^{d=6}$ and  $q=2,3,4$ and $7$.  Black curve at the bottom: lower bound on linear transformation. Blue curve (right on top of it): GLICA. Blue dashed curve: BloGLICA with $b=2$. Blue curve with X's: BloGLICA with $b=3$}
\label{fig4}
\end{figure}

Figure \ref{fig4} illustrates the performance of our suggested bound and algorithms when the alphabet size increases. Let $\underline{X} \in \{0,\dots, q-1\}^{d=6}$ be a six-dimensional random vector that follows a Zipf's law distribution with $s=1.01$. Here, we draw $10^6$ i.i.d. random samples of $\underline{X}$ for a dimension $d=6$ and different prime numbers $q$. We evaluate our suggested lower bound, as previously discussed. We further apply the GLICA algorithm and examine BloGLICA with $b=2$ and $3$. Figure \ref{fig4} demonstrates our objective for $q=2,3,5$ and $7$. 

\begin{figure}[h]
\centering
\includegraphics[width = 0.43\textwidth,bb= 40 193 550 590,clip]{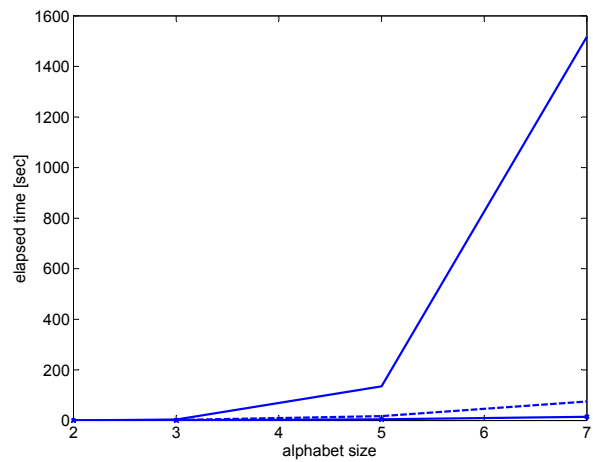}
\caption{Runtime of the experiment presented in Figure \ref{fig4}.}
\label{fig5}
\end{figure}

We first notice that GLICA achieves a sum of marginal entropies that is remarkably close the the lower bound. In addition, we notice that BloGLICA results in only a slight deterioration in performance (for $b=2,3$). As we examine the runtime of each of these methods (Figure \ref{fig5}), we notice a dramatic increase in GLICA's computational complexity when $q$ grows. As mentioned above, this computational drawback may be circumvented by applying the BloGLICA algorithm, at quite a mild overhead cost in the objective. 

We further compare our suggested approach to AMERICA, MEXICO and cobICA, under a generative linear mixture model assumption, as described in Section \ref{illustrations}. Here again, we observe that both GLICA and AMERICA successfully decompose the mixture model for smaller values of $d$ and $q$, while MEXICO and cobICA are slightly inferior. The detailed results are located in the first author's webpage\footnote{\url{https://sites.google.com/site/amichaipainsky/supplemental}}, due to space limitation.

\section{Applications}
Although ICA over finite fields does not originate from a specific application, it applies to a variety of problems. In \cite{gutch2012ica}, Gutch et al. suggested a specific application to this framework, in the context of eavesdropping on a multi-user MIMO digital communications system. Here, we show that ICA over finite fields also applies for large alphabet source coding. For the simplicity of the presentation we focus on the binary case.  

Consider $d$ independent binary sources with corresponding Bernoulli parameters $\{p_i\}_{i=1}^d$. Assume that the sources  are mixed by an unknown matrix $B$ over GF($2$). We draw $n$ i.i.d. samples from this mixture, which we would like to efficiently transmit to a receiver.  Large alphabet source coding considers the case where the number of draws $n$ is significantly smaller than the alphabet size $m=2^d$.  This means that the empirical entropy of drawn samples is significantly smaller the entropy of the source. Therefore, even if we assume that both the transmitter and the receiver know $\{p_i\}_{i=1}^d$, coding the samples according to the true distribution is quite wasteful. Large alphabet source coding has been extensively studies over the past decade. Several key contributions include theoretical performance bounds and practical methods in  different setups (see \cite{painsky2017large} for an overview). Here, we show that under a mixed independent sources assumption, our suggested binary ICA framework demonstrates favorable properties, both in terms of code-rate and run-time.  Specifically, we show that by applying our suggested GLICA scheme, we (strive) to recover the original independent sources. Then, we encode each recovered source independently, so that the sources are no longer considered over ``large alphabet" (as the alphabet is now binary). Notice that the redundancy of this scheme consists of three terms: the decomposition cost (\ref{eq:min_criterion}), the (negligible) redundancy of encoding the recovered binary sources, and the cost of describing the matrix $W$ to the receiver ($d^2$ bits).  

\begin{figure}[h]
\centering
\includegraphics[width = 0.49\textwidth,bb= 55 193 550 590,clip]{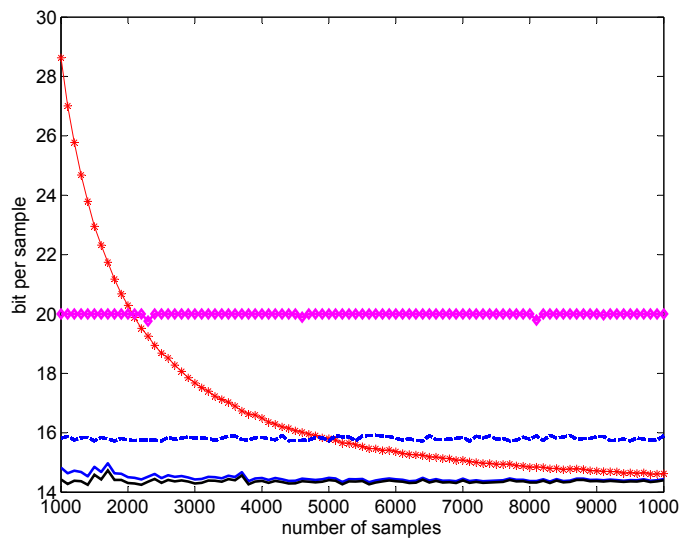}
\caption{Large alphabet source coding experiment. The red curve with the asterisks is a Huffman code according to the empirical joint distribution. The Blue curve is GLICA. The dashed blue curve is BloGLICA. The black curve is the sum of empirical marginal entropies of the original sources, and the magenta curve with the rhombuses is the sum of empirical marginal entropies of the mixed sources.}
\label{compression_experiment}
\end{figure}

We illustrate our suggested coding scheme in the following experiment. We draw $n$ i.i.d. samples from a random mixture of $d=20$ sources, with a corresponding set of parameters $\{p_i=\frac{i}{d}\}_{i=1}^{d}$. Notice that the joint entropy of this source is $14.36$ bits. We compare three different compression scheme. First, we examine the ``textbook" approach for this problem. Here, we construct a Huffman code for the $n$ samples, based on their empirical distribution. Since the receiver is oblivious to this code, we need to transmit it as well. This results in a total compression size of at least $n$ times the empirical entropy, plus a dictionary size of $n_0 \cdot d$, where $n_0$ is the number of unique symbols that appear in the $n$ samples. Second, we  apply our suggested GLICA algorithm, followed by arithmetic coding to each of the recovered sources. Finally, we apply BloGLICA with $b=2$, to reduce the run-time. Notice that the cost of describing the transformation is also reduced to $b \cdot \left(\frac{d}{b}\right)^2$.  Figure \ref{compression_experiment} demonstrated the compression rate we achieve for different values of $n$. The red curve with the asterisks corresponds to a Huffman compression (following \cite{witten1999managing}) with its corresponding dictionary. We first observe the notable effect of the dictionary's redundancy when $n<<2^d$, leading to a compression rate which is even greater than $d=20$ bits. However, as $n$ increases the relative portion of the dictionary decreases, since $n_0$ grows much slower. This leads to a quick decay in the compression rate. The blue curve at the bottom is GLICA, while the black curve is the empirical sum of marginal entropies of the original sources. Here we observe that GLICA successfully decomposes the sources, where the difference between the two curves is due to the cost of describing $W$ (which becomes negligible as $n$ increases). Further, we compare GLICA with marginal encoding to each of the mixed components  (that is, without trying to decompose it first). This results in an compression rate of approximately $20$ bits (magenta curve with rhombuses), as the mixture increased the empirical marginal entropies to almost a maximum. Finally, we apply the BloGLICA (blue dashed curve). We notice an increased compression rate compared with GLICA, due to a greater sum of marginal entropies. However, BloGLICA is a much more practical approach, as it takes only $25$ seconds to apply (for $n=10,000$), compared with $453$ seconds by GLICA. Notice that we omit other alternative methods that are inadequate or impractical to apply to this high dimensional problem  (cobICA, AMERICA, MEXICO).

To conclude, we show that by applying our suggested scheme we are able to efficiently compress data in a large alphabet regime. Although our results strongly depend on the independent sources  assumption, they are not limited to this model. In other words, we may apply GLICA to any set of  samples and  compare the resulting empirical sum of marginal entropies with the empirical joint entropy of the data; if the difference between the two is relatively small, then there is no significant loss in encoding the data component-wise, and the binary ICA compression scheme is indeed favorable. 

\section{On the flexibility of linear transformations}
\label{on_the_flexibility}
In the previous sections we introduced fundamental bounds and efficient algorithms for the linear ICA problem over finite fields. However, it is not quite clear how ``powerful" this tool is, for a given arbitrary mixture.  In other words, compared to its alternatives, how well can a linear transformation minimize the sum of marginal entropies, in the general case? To answer this question, we first need to specify an alternative approach. 

\subsection{Non-linear binary ICA}
\label{non-linear_BICA}
Assume we are interested in minimizing (\ref{eq:sum_ent_min_binary}) over non-linear invertible  transformations. As mentioned in Section \ref{previous_work}, this problem was originally introduced by Barlow \cite{barlow1989finding} as  minimally redundant representations, and is considered hard. 
In \cite{painsky2016generalized}, the authors introduce a piece-wise linear relaxation algorithm which tightly approximates the solution to (\ref{eq:sum_ent_min_binary}) with a series of linear problems. Here we focus on a simplified greedy algorithm which strives to minimize (\ref{eq:sum_ent_min_binary}) in a more computationally efficient manner with favorable theoretical properties. This approach, along with its computational and statistical characteristics,  was previously introduced in \cite{painsky2017large}. Here, we briefly review these results. 

\subsubsection{The Order Permutation}
\label{ordering solution}

As mentioned above, an invertible transformation $\underline{Y}=g(\underline{X})$ is a one-to-one mapping (i.e. permutation) of its $m=2^d$ alphabet symbols. In other words, an invertible transformation permutes the mapping of the $m$ symbols to the $m$ values of its probability mass function $\underline{p}=\{p_i\}_{i=1}^m$. Since our objective (\ref{eq:min_criterion}) is quite involved, we modify it by imposing an additional constraint. Specifically, we would like to sequentially minimize each term of the summation, $H(Y_j)$, for $j=1,\dots,d$, in a no-regret manner. As shown in \cite{painsky2017large}, the optimal solution to this problem is the order permutation, which suggests to map the $i^{th}$ smallest probability to the $i^{th}$ word (in its binary representation). This means, for example, that the all zeros word will be assigned to the smallest probability value in $\{p_i\}_{i=1}^m$ while the all ones word is assigned to the maximal probability value. 

\subsubsection{worst-case and average-case performance}
At this point, it is quite unclear how well the order permutation performs as a minimizer to (\ref{eq:min_criterion}). The following theorems present two theoretical properties which demonstrate its capabilities. These theorems (and their proofs) were first presented in \cite{painsky2017large}.

We begin with the worst-case performance of the order permutation. Here, we denote our objective (\ref{eq:min_criterion}) as $C(\underline{p},g)$,  as it solely depends on the probability vector and the transformation. Further, let us denote the order permutation as $g_{ord}$ and the optimal permutation (which minimizes (\ref{eq:min_criterion})) as  $g_{opt}$. 
In the worst-case analysis, we would like to quantify the maximum of $C(\underline{p},g_{opt})$ over all  probability mass functions $\underline{p}$, of a given alphabet size $m=2^d$. 

\begin{theorem}
\label{worst-case}
For any random vector $\underline{X}\sim \underline{p}$, over an alphabet size $m=2^d$ we have that
$$\max_{\underline{p}}C(\underline{p},g_{opt})=\Theta(d)$$
\end{theorem}
 Theorem \ref{worst-case} shows that even the optimal non-linear transformation achieves a sum of marginal entropies which is $\Theta(d)$ bits greater than the joint entropy, in the worst case. This means that there exists at least one source $\underline{X}$ with a probability mass function which is impossible to encode as if its components are independent without losing at least $\Theta(d)$ bits. This result is quite unfortunate since we have that $$C(\underline{p},g_{opt})=\sum_{j=1}^dH(Y_j)-H(\underline{Y})\leq \sum_{j=1}^dH(Y_j) \leq d.$$ Notice that this worst-case derivation obviously also applies for linear transformations. In other words, we can always find a worst-case source which no ICA algorithm (linear or non-linear) can efficiently decompose.

We now turn to an average-case analysis. Here, we show that the expected value of  $C(\underline{p},g_{ord})$ is bounded by a small constant, when averaging uniformly over all possible $\underline{p}$ over an alphabet size $m=2^d$. 
\begin{theorem}
\label{average case - asymp}
Let $\underline{X} \sim \underline{p}$ be a random vector of an alphabet size $m=2^d$ and joint probability mass function $\underline{p}$. Let $\underline{Y}=g_{ord}(\underline{X})$ be the order permutation. 
For $d \geq 10$, the expected value of $C(\underline{p},g_{ord})$, over a prior uniform simplex of joint probability mass functions $\underline{p}$, satisfies
\begin{align}
\label{block_perm}
\mathbb{E}_{\underline{\smash{p}}}C(\underline{p},g_{ord}) <0.0162 +O\left(\frac{1}{m}\right).
\end{align}
\end{theorem}
\noindent This means that when the alphabet size is large enough, the order permutation achieves, on the average, a sum of marginal entropies which is only $0.0162$ bits greater than the joint entropy, when all possible probability mass functions $\underline{p}$ are equally likely to appear. Notice that the uniform prior assumption may not be adequate for every setup, but it does provide a universal result for the performance of this minimizer.  Proofs of these theorems are located in \cite{painsky2017large} under different notations, and in the first author's webpage\footnote{\url{https://sites.google.com/site/amichaipainsky/supplemental}} in the same notation that is used above.

\subsection{Average-case performance of Linear ICA transformations}
\label{Average-case performance of Linear ICA transformations}
As with the non-linear case, we would like to evaluate the average-case performance of linear transformations. 
\begin{theorem}
\label{average case - linear}
Let $\underline{X} \sim \underline{p}$ be a random vector of an alphabet size $m=2^d$ and joint probability mass function $\underline{p}$. Let  $\underline{Y}^*=W_{\underline{p}}^*(\underline{X})$ be the optimal linear transformation (notice that the optimal transformation $W_{\underline{p}}^*$ depends on $\underline{p}$). Then, the expected value of $\sum_{j=1}^d H(Y^*_j)$, over a prior uniform simplex of joint probability mass functions $\underline{p}$, satisfies
\begin{align}
\lim_{d \rightarrow \infty} \frac{1}{d}\mathbb{E}_{\underline{\smash{p}}}\sum_{j=1}^d H(Y^*_j)=1 
\end{align}
\end{theorem}
A proof for this theorem is provided in the Appendix. This theorem suggests that when the number of components is large enough, even the optimal linear transformation preforms very poorly on the average, as it attains the maximal possible marginal entropy value ($E(Y_j)\leq1$). Moreover, it is shown in the Appendix that this average-case performance is equivalent to applying no transformation at all. In other words, when the number of components is too large, linear transformations are useless in the worst-case, and on the average. The intuition behind these results is that our objective depends on the entire probability mass function of $\underline{X}$ (which is of order $2^d$), while the number of free parameters, when applying linear transformations, is only $O\left(d^2\right)$. This means that when $d$ increases, linear transformations are simply not flexible enough to minimize the objective. 

\subsection{Illustrations}
\label{illustrations2}
\begin{figure}[h]
\centering
\includegraphics[width = 0.42\textwidth,bb= 50 193 550 590,clip]{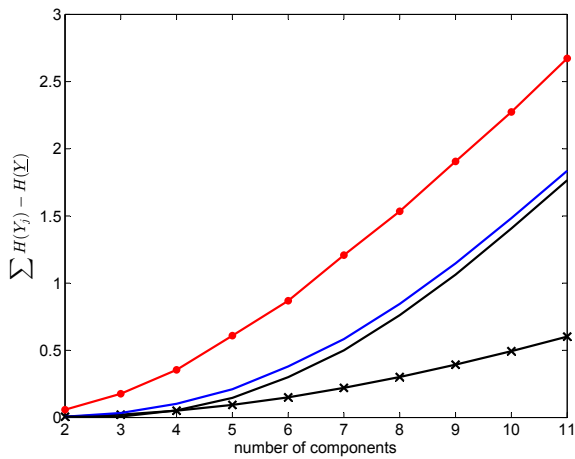}
\caption{Minimizing (\ref{eq:sum_ent_min_binary}) given independent draws from a Zipf's distribution, using linear and non-linear methods. Black curve with X's: the order permutation.  Black curve: lower bound on linear transformation. Blue curve above it: GLICA. Red curve with circles: applying no transformations.}
\label{fig6}
\end{figure}

Let us now compare our suggested linear algorithms and bound with the non-linear order permutation discussed above. In the first experiment we draw $10^6$ independent samples from a Zipf law distribution with $s=1.01$ and a varying number of components $d$. We evaluate the lower bound of (\ref{eq:sum_ent_min_binary}) under linear transformations, as discussed in Section \ref{linear_BICA_lowerbound} and further apply the GLICA algorithm (Section \ref{linear_BICA_algo}). We compare our linear results with the non-linear order permutation on one hand, and with applying no transformation at all on the other hand. Figure \ref{fig6} demonstrates the results we achieve. As we can see, the order permutation outperforms any linear solution quite remarkably, for the reasons mentioned above. Moreover, we notice that the gap increases as the number of components grows.  

\begin{figure}[h]
\centering
\includegraphics[width = 0.42\textwidth,bb= 50 193 550 590,clip]{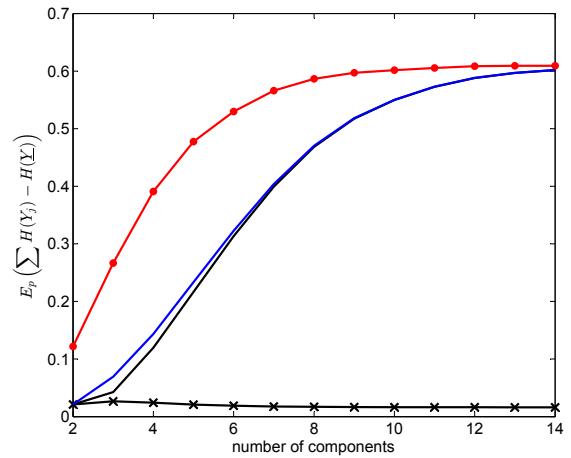}
\caption{Average-case analysis of minimizing (\ref{eq:sum_ent_min_binary})  using linear and non-linear methods. The curves are described in Figure \ref{fig6} }
\label{fig7}
\end{figure}

We now turn to illustrate the average-case performance, as discussed in previous sections. Figure \ref{fig7} shows the mean of our objective (empirically evaluated over $10^7$ independent draws from a uniform simplex), for the four methods mentioned above. We first see that the non-linear order permutation converges to a small overhead constant, as indicated in Theorem \ref{average case - asymp}. On the other hand, the lower bound of the linear solution behaves asymptotically like applying no transformation at all. This can be easily derived from Theorem \ref{average case - linear} and the fact that  $\mathbb{E}_{\underline{\smash{p}}}H(\underline{X})=\frac{1}{log_e(2)}\left(\Psi(2^d+1)-\Psi(2)\right))$, where $\Psi(\cdot)$ is a digamma function, as shown in \cite{painsky2017large}. 

\section{Conclusion}
In this work we consider a framework for ICA over finite fields, in which we strive to decompose any given  vector into ``as statistically independent as possible" components, using only linear transformations. 
Over the years, several solutions have been proposed to this problem. Most of them strongly depend on the assumption that a perfect decomposition exists, while the rest suggest a variety of heuristics to the general case.  In this work, we present a novel lower bound which provides a fundamental limit for the performance of any linear solution. Based on this bound, we suggest a simple greedy algorithm which shows  favorable statistically properties, compared with our suggested bound. In addition, we introduced a simple modification to our suggested algorithm which significantly reduces its computational complexity at a relatively small cost in the objective.

In addition, we discuss the basic limitations of working with linear transformations. We show that this class of  transformations becomes quite ineffective in the general case,  when the dimension of the problem increases. Specifically, we show that when averaging over a uniform prior, applying the optimal linear transformation is equivalent to applying no transformation at all. This result should not come as a surprise, since the objective depends on the full statistical description of the problem (which is exponential in the number of components $d$), while linear transformations only provides $O\left(d^2\right)$ free parameters. That being said, we do not intend to discourage the use of linear transformations for finite field ICA. Our analysis focuses on a universal setup in which no prior assumption is made. This is not always the case in real-world applications. For example, linear transformations may be quite effective in cases where we assume a generative linear model but the sources are not completely independent. 

Although the focus of this paper is mostly theoretical, linear ICA  is shown to have many applications in a variety of fields. We believe that the theoretical properties that we introduce, together with the practical algorithms which utilize a variety of setups, make our contribution applicable to many disciplines. 

\appendix 
This Appendix provide the proof of Theorem \ref{average case - linear}. We begin the proof by introducing the following Lemma:
\begin{appendix_lemma}
\label{lemma}
Let $\underline{Y} \sim \underline{p}$ be a binary random vector of $d$ components, $\underline{Y} \in \{0,1\}^d$. Then, 
$$\mathbb{E}_{\underline{\smash{p}}}\sum_{j=1}^d H(Y_j)=\frac{d}{\log_e(2)}\left(\Psi(2^d-1)-\Psi(2^{d-1}) \right)$$
where the expectation is over a prior of uniform simplex of probability mass functions $\underline{p}$ and $\psi(\cdot)$ is the digamma function. 
\end{appendix_lemma}
\begin{proof}
The probability vector $\underline{p}$ consists of $2^d$ elements and follows a uniform simplex prior. This means that it follows a flat Dirichlet distribution with a parameter $\alpha=\frac{1}{m}$. The marginal probability of each component of $\underline{Y}$ is a summation of half of the elements of $\underline{p}$. 
Therefore, following the Dirichlet distribution properties, we have that $p_y   \triangleq  p(y=0)$ again follows a Dirichlet distribution, with $\alpha=2^{d-1}$. This means that $f(p_y)  = \frac{\Gamma(2^d)}{\Gamma(2^{d-1})^2} p_y^{2^{d-1}-1}(1-p_y)^{2^{d-1}-1}=\text{Beta}(2^{d-1}-1,2^{d-1}-1)$. Notice that for a symmetric Beta distributed random variable $X\sim Beta(\alpha,\alpha)$, we have that $\mathbb{E}(X\log_e (X))=\frac{1}{2}\left(\Psi(\alpha+1) - \Psi(2\alpha+1)\right)$ where $\psi(\cdot)$ is the digamma function.  Therefore, 
\begin{align}
&\mathbb{E}_{\underline{\smash{p}}}\sum_{j=1}^d H(Y_j)=\\\nonumber
&d\int\left(-p_y\log p_y-(1-p_y)\log(1-p_y) \right)f(p_y)dp_y=\\\nonumber
&\frac{-2d}{\log_e(2)}\mathbb{E}(X\log_e (X))=\frac{d}{\log_e(2)}\left(\Psi(2^d-1)-\Psi(2^{d-1}) \right)
\end{align} 
\begin{flushright}$\square$\end{flushright}
\end{proof}

\noindent\textbf{Proof of Theorem \ref{average case - linear}:}
Let $\underline{Y} \sim \underline{p}$ be a binary random vector of $d$ components, $\underline{Y} \in \{0,1\}^d$. For the simplicity of the presentation, denote our objective as $\sum_{j=1}^d H(Y_j) \triangleq  L(\underline{p})$. Its expectation over a prior of uniform simplex of probability mass functions is defined as $\mathbb{E}_{\underline{\smash{p}}}L(\underline{p})=C\int_S L(\underline{p})d\underline{p}$ where $S=\left\{\underline{p}\; | \;p_i \in [0,1], \;\; \sum p_i=1  \right\}$ is the unit simplex and $C$ is a normalization constant such that $ C\int_{S}d\underline{p}=1$. Let us use the symmetry of the simplex to reformulate this definition of expectation. Denote
$$S_u=\left\{\underline{q}\; | \;q_i \in [0,1], \;\; \sum q_i=1, \;\;  q_1\leq q_{2}\leq \dots \leq q_{2^d} \right\}, $$
$$ R(\underline{q})=\left\{\underline{p}\; | \;\underline{p}\; \text{is a permutation of}\; \underline{q} \right\},$$
where $S_u$ is the set of all ascending ordered probability vectors of size $2^d$ and $R(\underline{q})$ is the set of all possible permutations of the vector $\underline{q}$. 
Notice we have that $\underline{p} \in S$ iff $\exists \; \underline{q}$ such that $\underline{p} \in R(\underline{q})$ and $\underline{q}\in S_u$. In other words, we can express every probability mass function as a permutation of an ascending ordered probability mass function.  Therefore, $\mathbb{E}_{\underline{\smash{p}}}L(\underline{p})=C\int_{\underline{q}\in S_u} \sum_{\underline{p} \in R(\underline{q})} L(\underline{p})d\underline{q}$.

Let us now focus on the set $R(\underline{q})$, for a fixed $\underline{q}$. Notice that this set represents the set of all invertible transformations of $\underline{X} \sim \underline{q}$. In other words, assume $\underline{X} \sim \underline{q}$, then $\underline{Y}=g(\underline{X})$ is an invertible transformation iff $\underline{Y} \sim \underline{p} \in R(\underline{q})$. Linear invertible transformations define a (disjoint) partitioning of $R(\underline{q})$. This means that for every $Y \sim \underline{p} \in R(\underline{q})$, we can define an invertible linear transformation $\underline{\tilde{Y}}=W\underline{Y}$ such that $\underline{\tilde{Y}} \sim \underline{\tilde{p}} \in R_w(\underline{p},\underline{q}) \subset R(\underline{q})$. In other words, for every $\underline{p} \in R(\underline{q})$ we can define a set of probability mass functions $R_w(\underline{p},\underline{q})$ that are a result of all invertible linear transformations on it. These sets are disjoint since every set of linear transformations is closed; if  $p_0 \in R_w(\underline{p}_1,\underline{q})$ and $p_0 \in R_w(\underline{p}_2,\underline{q})$, then there exists a linear transformation from any element in $R_w(\underline{p}_1,\underline{q})$ to any element in $R_w(\underline{p}_2,\underline{q})$, which contradicts the definition of $R_w(\underline{p},\underline{q})$ above. Since these transformations are invertible (by definition), they are also elements in $R(\underline{q})$. Denote the size of the set $R_w(\underline{p},\underline{q}) $ as $A$. Notice that $|R(\underline{q})|=2^d!$ while $|R_w(\underline{p},\underline{q})|=\frac{\prod_{l=0}^{d-1}2^d-2^l}{d!}=O\left(2^{d^2}\right)$, as shown in \cite{gutch2012ica}. Notice that these figures only depend of the dimension of the problem (and not on $\underline{p}$ or $\underline{q}$).   

In every set $R_w(\underline{p},\underline{q})$ there exists (at least one) minimizer of our objective. We denote it by $\underline{p}^*_w$. Further, define the set of all linear minimizers as $R^*(\underline{q})$. 
We have that 
\begin{align}
\sum_{\underline{p}\in R(\underline{q})}L(\underline{p})=&\sum_{p\in R^*(\underline{q})}L(\underline{p})+\sum_{p\in R(\underline{q}) \setminus R^*(\underline{q})}L(\underline{p}) \leq \\\nonumber
&\sum_{p\in R^*(\underline{q})}L(\underline{p})+ d|R(\underline{q}) \setminus R^*(\underline{q})|
\end{align}
This means that 
\begin{align}
\sum_{p\in R^*(\underline{q})}L(\underline{p}) \geq \sum_{\underline{p}\in R(\underline{q})}L(\underline{p})-d\left(|R(\underline{q})|-\frac{|R(\underline{q})|}{A}\right)
\end{align}
since $|R^*(\underline{q})|={|R(\underline{q})|}/{A}$. Let us now multiply both sides by $A$ and average over all $\underline{q}$. We get that
\begin{align}
&C\int_{\underline{q}\in S_u} A \sum_{\underline{p} \in R^*(\underline{q})} L(\underline{p})d\underline{q}\geq \\\nonumber
& A \cdot  C \int_{\underline{q}\in S_u} \left(  \sum_{\underline{p} \in R(\underline{q})} L(\underline{p}) -  d\left(|R(\underline{q})|-\frac{|R(\underline{q})|}{A}\right)\right)d\underline{q}=\\\nonumber
&A\mathbb{E}_{\underline{\smash{p}}}L(\underline{p})-A\cdot d \left(1-\frac{1}{A}\right)=A\mathbb{E}_{\underline{\smash{p}}}L(\underline{p})-A\cdot d +d
\end{align}
where the equality follows from $ C\int_{\underline{q}\in S_u}|R(\underline{q})|d\underline{q}=1$. Finally, we have that 
\begin{align}
\lim_{d\rightarrow \infty}\frac{1}{d}C\int_{\underline{q}\in S_u} &A \sum_{\underline{p} \in R^*(\underline{q})} L(\underline{p})d\underline{q}\geq\\\nonumber
& A \lim_{d\rightarrow \infty} \frac{1}{d}\mathbb{E}_{\underline{\smash{p}}}L(\underline{p})-A+1=1
\end{align}
due to Lemma \ref{lemma} above.  Notice that the expression on the left is exactly the expected value of the optimal linear solution, as desired. 
\begin{flushright}$\square$\end{flushright}
\balance
\bibliographystyle{IEEEtran}
\bibliography{bibi}

\begin{thebibliography}{10}
\providecommand{\url}[1]{#1}
\csname url@samestyle\endcsname
\providecommand{\newblock}{\relax}
\providecommand{\bibinfo}[2]{#2}
\providecommand{\BIBentrySTDinterwordspacing}{\spaceskip=0pt\relax}
\providecommand{\BIBentryALTinterwordstretchfactor}{4}
\providecommand{\BIBentryALTinterwordspacing}{\spaceskip=\fontdimen2\font plus
\BIBentryALTinterwordstretchfactor\fontdimen3\font minus
  \fontdimen4\font\relax}
\providecommand{\BIBforeignlanguage}[2]{{%
\expandafter\ifx\csname l@#1\endcsname\relax
\typeout{** WARNING: IEEEtran.bst: No hyphenation pattern has been}%
\typeout{** loaded for the language `#1'. Using the pattern for}%
\typeout{** the default language instead.}%
\else
\language=\csname l@#1\endcsname
\fi
#2}}
\providecommand{\BIBdecl}{\relax}
\BIBdecl

\bibitem{barlow1989finding}
H.~Barlow, T.~Kaushal, and G.~Mitchison, ``Finding minimum entropy codes,''
  \emph{Neural Computation}, vol.~1, no.~3, pp. 412--423, 1989.

\bibitem{schmidhuber1992learning}
J.~Schmidhuber, ``Learning factorial codes by predictability minimization,''
  \emph{Neural Computation}, vol.~4, no.~6, pp. 863--879, 1992.

\bibitem{painsky2016generalized}
A.~Painsky, S.~Rosset, and M.~Feder, ``Generalized independent component
  analysis over finite alphabets,'' \emph{IEEE Transactions on Information
  Theory}, vol.~62, no.~2, pp. 1038--1053, 2016.

\bibitem{becker1988improving}
S.~Becker and Y.~Le~Cun, ``Improving the convergence of back-propagation
  learning with second order methods,'' in \emph{Proceedings of the 1988
  connectionist models summer school}.\hskip 1em plus 0.5em minus 0.4em\relax
  San Matteo, CA: Morgan Kaufmann, 1988, pp. 29--37.

\bibitem{becker1996unsupervised}
S.~Becker and M.~Plumbley, ``Unsupervised neural network learning procedures
  for feature extraction and classification,'' \emph{Applied Intelligence},
  vol.~6, no.~3, pp. 185--203, 1996.

\bibitem{obradovic1996information}
D.~Obradovic, \emph{An information-theoretic approach to neural
  computing}.\hskip 1em plus 0.5em minus 0.4em\relax Springer Science \&
  Business Media, 1996.

\bibitem{choi2000factorial}
S.~Choi and O.~Lee, ``Factorial code representation of faces for recognition,''
  in \emph{Biologically Motivated Computer Vision}.\hskip 1em plus 0.5em minus
  0.4em\relax Springer, 2000, pp. 42--51.

\bibitem{bartlett2002face}
M.~S. Bartlett, J.~R. Movellan, and T.~J. Sejnowski, ``Face recognition by
  independent component analysis,'' \emph{IEEE Transactions on Neural
  Networks}, vol.~13, no.~6, pp. 1450--1464, 2002.

\bibitem{bartlett2007information}
M.~S. Bartlett, ``Information maximization in face processing,''
  \emph{Neurocomputing}, vol.~70, no.~13, pp. 2204--2217, 2007.

\bibitem{schmidhuber2011fast}
J.~Schmidhuber, D.~Cire{\c{s}}an, U.~Meier, J.~Masci, and A.~Graves, ``On fast
  deep nets for agi vision,'' in \emph{Artificial General Intelligence}.\hskip
  1em plus 0.5em minus 0.4em\relax Springer, 2011, pp. 243--246.

\bibitem{schmidhuber2015deep}
J.~Schmidhuber, ``Deep learning in neural networks: An overview,'' \emph{Neural
  Networks}, vol.~61, pp. 85--117, 2015.

\bibitem{painsky2015universal}
A.~Painsky, S.~Rosset, and M.~Feder, ``Universal compression of memoryless
  sources over large alphabets via independent component analysis,'' in
  \emph{Data Compression Conference (DCC)}.\hskip 1em plus 0.5em minus
  0.4em\relax IEEE, 2015, pp. 213--222.

\bibitem{painsky2016simple}
------, ``A simple and efficient approach for adaptive entropy coding over
  large alphabets,'' in \emph{Data Compression Conference (DCC)}.\hskip 1em
  plus 0.5em minus 0.4em\relax IEEE, 2016, pp. 369--378.

\bibitem{painsky2017large}
------, ``Large alphabet source coding using independent component analysis,''
  \emph{IEEE Transactions on Information Theory}, vol.~63, no.~10, pp.
  6514--6529, 2017.

\bibitem{painsky2018}
A.~Painsky, ``Phd dissertation: {G}eneralized {I}ndependent {C}omponents
  {A}nalysis over finite alphabets,'' \emph{arXiv preprint arXiv:1809.05043},
  2018.

\bibitem{yeredor2007ica}
A.~Yeredor, ``I{C}{A} in boolean {XOR} mixtures,'' in \emph{Independent
  Component Analysis and Signal Separation}.\hskip 1em plus 0.5em minus
  0.4em\relax Springer, 2007, pp. 827--835.

\bibitem{yeredor2011independent}
------, ``Independent analysis over {G}alois fields of prime order,''
  \emph{IEEE Transactions on Information Theory}, vol.~57, no.~8, pp.
  5342--5359, 2011.

\bibitem{vsingliar2006noisy}
T.~{\v{S}}ingliar and M.~Hauskrecht, ``Noisy-or component analysis and its
  application to link analysis,'' \emph{The Journal of Machine Learning
  Research}, vol.~7, pp. 2189--2213, 2006.

\bibitem{streich2009multi}
A.~P. Streich, M.~Frank, D.~Basin, and J.~M. Buhmann, ``Multi-assignment
  clustering for {B}oolean data,'' in \emph{Proceedings of the 26th Annual
  International Conference on Machine Learning}.\hskip 1em plus 0.5em minus
  0.4em\relax ACM, 2009, pp. 969--976.

\bibitem{nguyen2011binary}
H.~Nguyen and R.~Zheng, ``Binary {I}ndependent {C}omponent {A}nalysis with {OR}
  mixtures,'' \emph{IEEE Transactions on Signal Processing}, vol.~59, no.~7,
  pp. 3168--3181, 2011.

\bibitem{attux2011immune}
D.~G. Silva, R.~Attux, E.~Z. Nadalin, L.~T. Duarte, and R.~Suyama, ``An
  immune-inspired information-theoretic approach to the problem of {ICA} over a
  galois field,'' in \emph{Information Theory Workshop (ITW)}.\hskip 1em plus
  0.5em minus 0.4em\relax IEEE, 2011, pp. 618--622.

\bibitem{wood2012non}
F.~Wood, T.~Griffiths, and Z.~Ghahramani, ``A non-parametric {B}ayesian method
  for inferring hidden causes,'' \emph{arXiv preprint arXiv:1206.6865}, 2012.

\bibitem{painsky2016binary}
A.~Painsky, S.~Rosset, and M.~Feder, ``Binary independent component analysis:
  Theory, bounds and algorithms,'' in \emph{The International Workshop on
  Machine Learning for Signal Processing (MLSP)}.\hskip 1em plus 0.5em minus
  0.4em\relax IEEE, 2016, pp. 1--6.

\bibitem{atick1990towards}
J.~J. Atick and A.~N. Redlich, ``Towards a theory of early visual processing,''
  \emph{Neural Computation}, vol.~2, no.~3, pp. 308--320, 1990.

\bibitem{plumbley1993efficient}
M.~D. Plumbley, ``Efficient information transfer and anti-hebbian neural
  networks,'' \emph{Neural Networks}, vol.~6, no.~6, pp. 823--833, 1993.

\bibitem{gutch2012ica}
H.~W. Gutch, P.~Gruber, A.~Yeredor, and F.~J. Theis, ``I{CA} over finite fields
  - separability and algorithms,'' \emph{Signal Processing}, vol.~92, no.~8,
  pp. 1796--1808, 2012.

\bibitem{silva2014cobica}
D.~G. Silva, J.~Montalvao, and R.~Attux, ``cob{ICA}: A concentration-based,
  immune-inspired algorithm for ica over galois fields,'' in \emph{IEEE
  Symposium on Computational Intelligence for Multimedia, Signal and Vision
  Processing (CIMSIVP)}, 2014, pp. 1--8.

\bibitem{watanabe1960information}
S.~Watanabe, ``Information theoretical analysis of multivariate correlation,''
  \emph{IBM Journal of research and development}, vol.~4, no.~1, pp. 66--82,
  1960.

\bibitem{painsky2014generalized}
A.~Painsky, S.~Rosset, and M.~Feder, ``Generalized binary independent component
  analysis,'' in \emph{International Symposium on Information Theory
  (ISIT)}.\hskip 1em plus 0.5em minus 0.4em\relax IEEE, 2014, pp. 1326--1330.

\bibitem{mulmuley1986fast}
K.~Mulmuley, ``A fast parallel algorithm to compute the rank of a matrix over
  an arbitrary field,'' in \emph{Proceedings of the eighteenth annual ACM
  symposium on Theory of computing}.\hskip 1em plus 0.5em minus 0.4em\relax
  ACM, 1986, pp. 338--339.

\bibitem{shulman2003communication}
N.~Shulman, ``Communication over an unknown channel via common broadcasting,''
  Ph.D. dissertation, 2003.

\bibitem{witten1999managing}
I.~H. Witten, A.~Moffat, and T.~C. Bell, \emph{Managing gigabytes: compressing
  and indexing documents and images}.\hskip 1em plus 0.5em minus 0.4em\relax
  Morgan Kaufmann, 1999.

\end{thebibliography}


\begin{IEEEbiography}[{\includegraphics[width=1in,height=1.25in,clip,keepaspectratio]{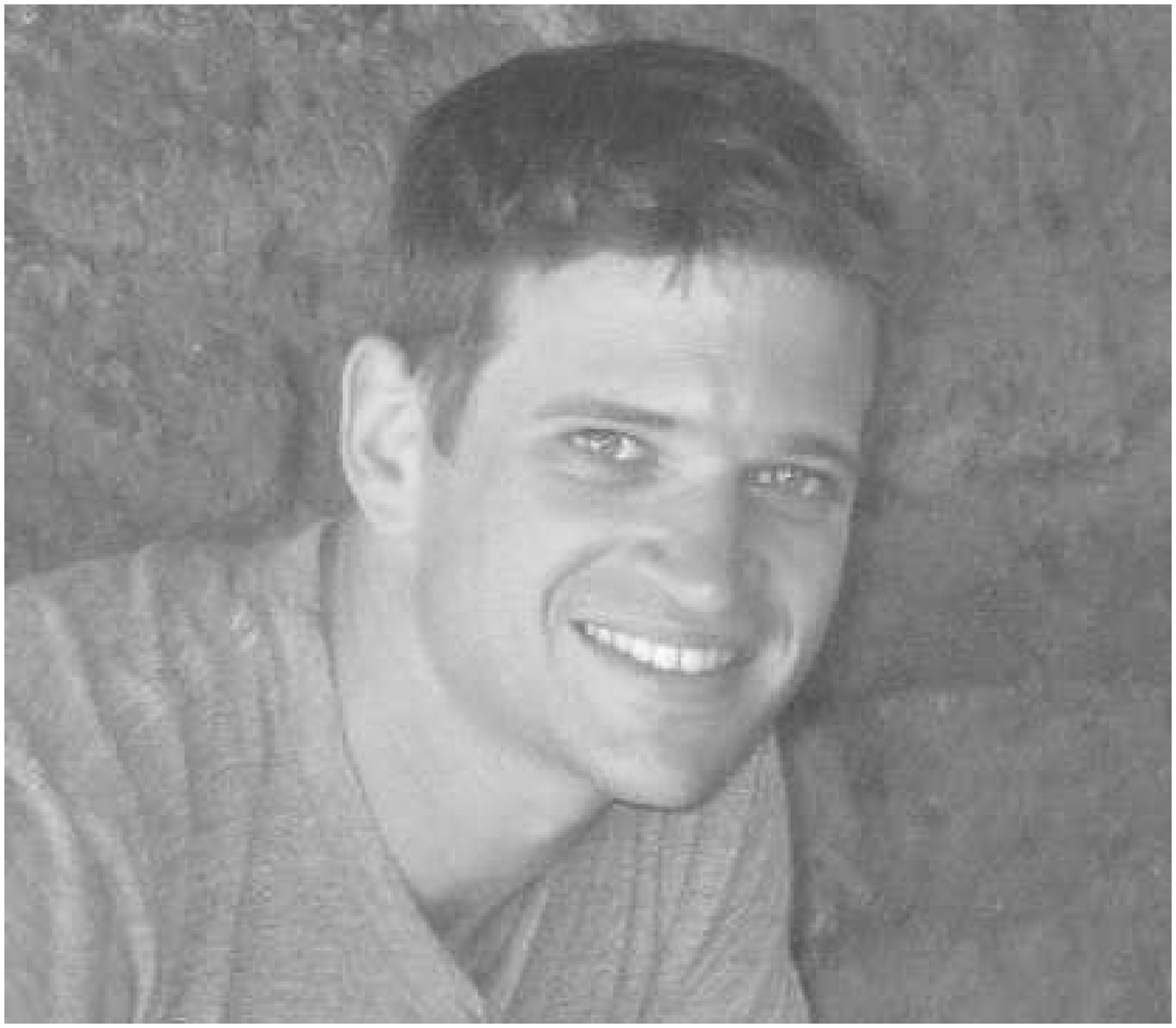}}]{Amichai Painsky}
received his B.Sc. in Electrical Engineering from Tel Aviv University (2007), his M.Eng. degree in Electrical Engineering from Princeton University (2009) and his Ph.D. in Statistics from the  School of Mathematical Sciences in Tel Aviv University. He is currently a Post-Doctoral Fellow, co-affiliated with the
Israeli Center of Research Excellence in Algorithms (I-CORE) at
the Hebrew University of Jerusalem, and the Signals, Information and Algorithms
(SIA) Lab at MIT. His research interests include Data Mining, Machine Learning, Statistical Learning and their connection to Information Theory.
\end{IEEEbiography}

\begin{IEEEbiography}[{\includegraphics[width=1in,height=1.25in,clip,keepaspectratio]{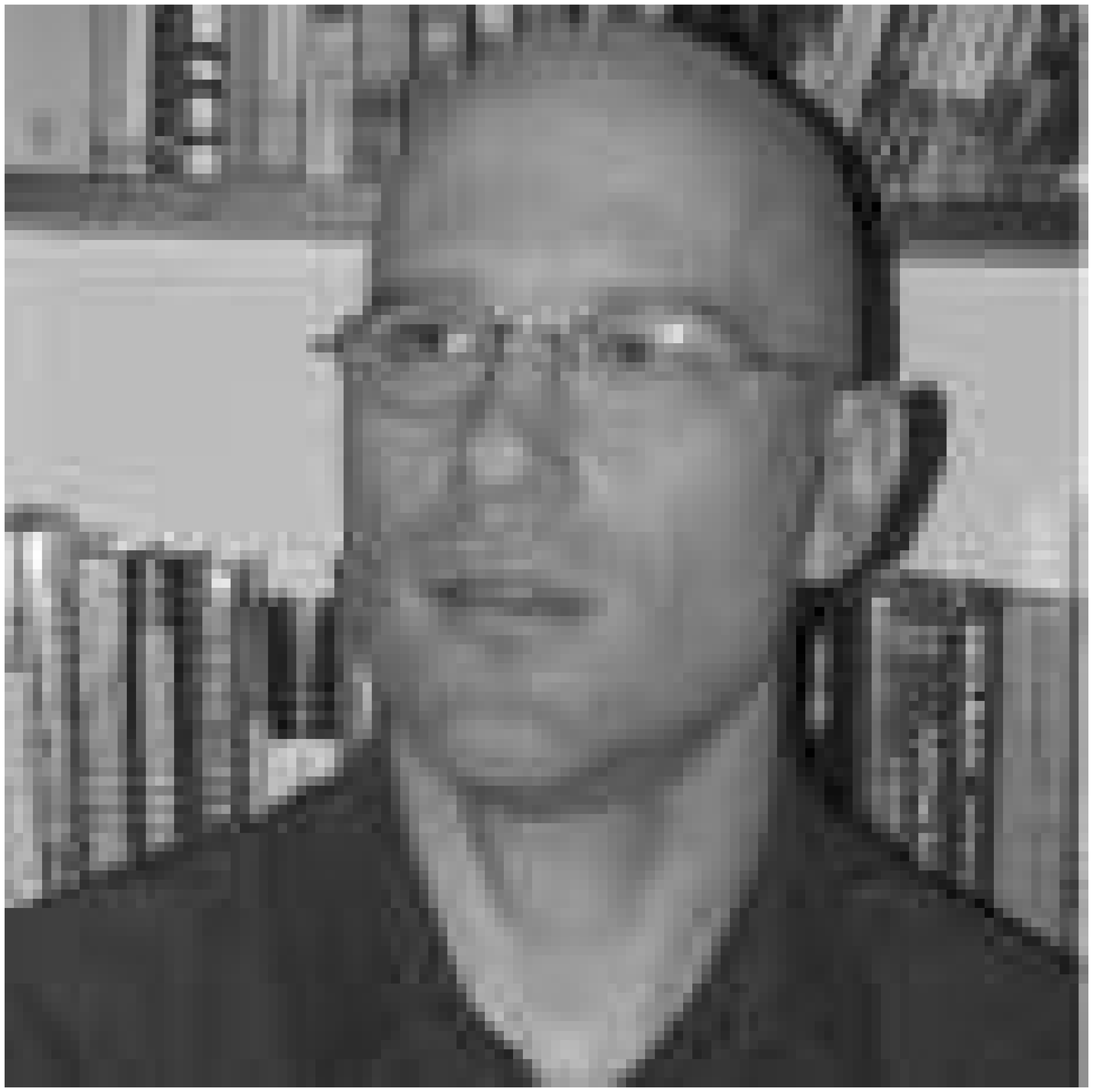}}]{Saharon Rosset}
is a Professor in the department of Statistics and Operations Research at Tel Aviv University. His research interests are in Computational Biology and Statistical Genetics, Data Mining and Statistical Learning. Prior to his tenure at Tel Aviv, he received his PhD from Stanford University in 2003 and spent four years as a Research Staff Member at IBM Research in New York. He is a five-time winner of major data mining competitions, including KDD Cup (four times) and INFORMS Data Mining Challenge, and two time winner of the best paper award at KDD (ACM SIGKDD International Conference on Knowledge Discovery and Data Mining)
\end{IEEEbiography}

\begin{IEEEbiography}[{\includegraphics[width=0.8in,height=1in,clip,keepaspectratio]{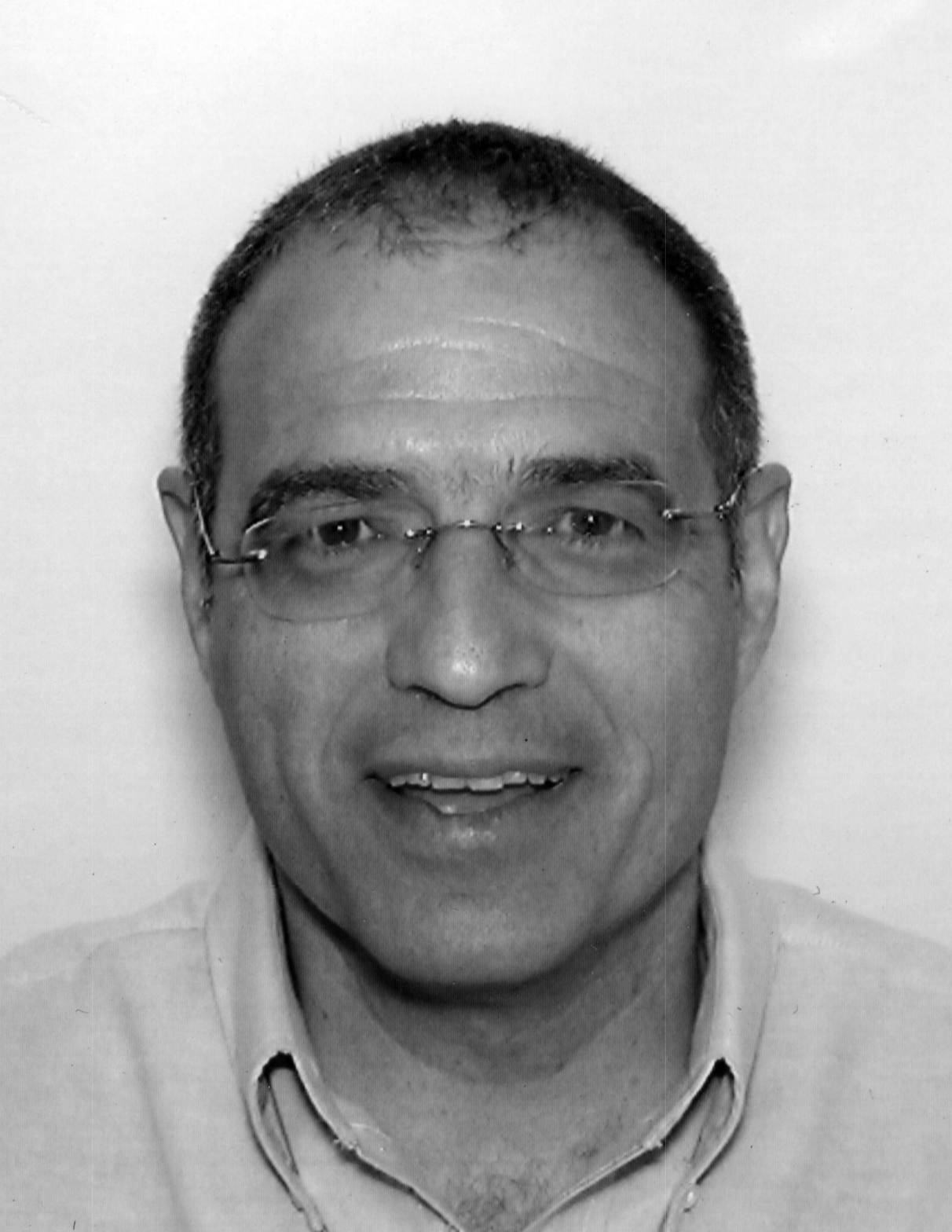}}]{Meir Feder}
(S'81-M'87-SM'93-F'99) received the B.Sc and M.Sc degrees
from Tel-Aviv University, Israel and the Sc.D degree from the Massachusetts
Institute of Technology (MIT) Cambridge, and the Woods Hole
Oceanographic Institution, Woods Hole, MA, all in electrical
engineering in 1980, 1984 and 1987, respectively.

After being a research associate and lecturer in MIT he joined
the Department of Electrical Engineering - Systems, School of Electrical Engineering, Tel-Aviv
University, where he is now a Professor and the incumbent of the Information Theory Chair.
He had visiting appointments at the Woods Hole Oceanographic Institution, Scripps
Institute, Bell laboratories and has been a visiting
professor at MIT. He is also extensively involved in the high-tech
industry as an entrepreneur and angel investor.
He co-founded several companies including Peach Networks,
a developer of a server-based interactive TV solution which
was acquired by Microsoft, and Amimon a
provider of ASIC's for wireless high-definition A/V connectivity.
Prof. Feder is a co-recipient of the 1993 IEEE Information Theory
Best Paper Award. He also received the 1978 "creative thinking"
award of the Israeli Defense Forces, the 1994 Tel-Aviv University
prize for Excellent Young Scientists, the 1995 Research Prize of
the Israeli Electronic Industry, and the research prize in applied
electronics of the Ex-Serviceman Association, London, awarded by
Ben-Gurion University.
\end{IEEEbiography}

\vfill



\end{document}